\documentclass[sigplan]{acmart}
\usepackage{multirow}

\usepackage{epsfig}
\usepackage{graphicx}
\usepackage[switch]{lineno}
\usepackage{bm}
 
\usepackage{amssymb}
\usepackage{amsmath}

\begin{document}
\title{Feature Mining: A Novel Training Strategy for Convolutional Neural Network}
\author{\textbf{Tianshu Xie}\textsuperscript{1}}
\orcid{1234-4564-1234-4565}
\affiliation{%
\streetaddress{Samuels Building (F25), Kensington Campus}
\postcode{2052}
\country{tianshuxie@std.uestc.edu.cn}}
\author{\textbf{Xuan Cheng}\textsuperscript{1}}
\orcid{1234-4564-1234-4565}
\affiliation{%
\streetaddress{Samuels Building (F25), Kensiton Campus}
\postcode{2052}
\country{cs\_xuancheng@std.uestc.edu.cn}}
\author{\textbf{Xiaomin Wang}\textsuperscript{1}}
\orcid{1234-4565-4564-1234}
\affiliation{%
\country{xmwang@uestc.edu.cn}}
\author{\textbf{Minghui Liu}\textsuperscript{1}}
\orcid{1234-4564-1234-4565}
\affiliation{%
\country{minghuiliuuestc@163.com}}
\author{\textbf{Jiali Deng}\textsuperscript{1}}
\affiliation{%
\country{julia\_d@163.com}}
\author{\textbf{Ming Liu}\textsuperscript{1*}}

\orcid{1234-4565-4564-1234}
\affiliation{%
\country{csmliu@uestc.edu.cn\\}}\par
\par
\par
\bigskip

\begin{abstract}
In this paper, we propose a novel training strategy for convolutional neural network(CNN) named Feature Mining, that aims to strengthen the network's learning of the local feature. Through experiments, we find that semantic contained in different parts of the feature is different, while the network will inevitably lose the local information during feedforward propagation. In order to enhance the learning of local feature, Feature Mining divides the complete feature into two complementary parts and reuse these divided feature to make the network learn more local information, we call the two steps as feature segmentation and feature reusing. Feature Mining is a parameter-free method and has plug-and-play nature, and can be applied to any CNN models. Extensive experiments demonstrate the wide applicability, versatility, and compatibility of our method.

\end{abstract}


\maketitle

\section{Introduction}
Convolution neural network (CNN) has made significant progress on various computer vision tasks, $e.g$, image classification~\cite{krizhevsky2012imagenet,russakovsky2015imagenet,szegedy2015going,he2016deep}, object detection~\cite{girshick2015fast,ren2015faster,he2017mask}, and segmentation~\cite{long2015fully,chen2017deeplab}. However, the large scale and tremendous parameters of CNN may incur overfitting and reduce generalizations, that bring challenges to the network training. A series of training strategies proposed to solve these problems, including data augmentation~\cite{devries2017improved,zhong2020random,zhang2017mixup}, batch normalization~\cite{ioffe2015batch} and knowledge distillation~\cite{hinton2015distilling}. Those approaches make the network training more robust from the perspective of input, feature space and output.

\begin{figure}[t]
\begin{center}
\includegraphics[width=0.95\linewidth]{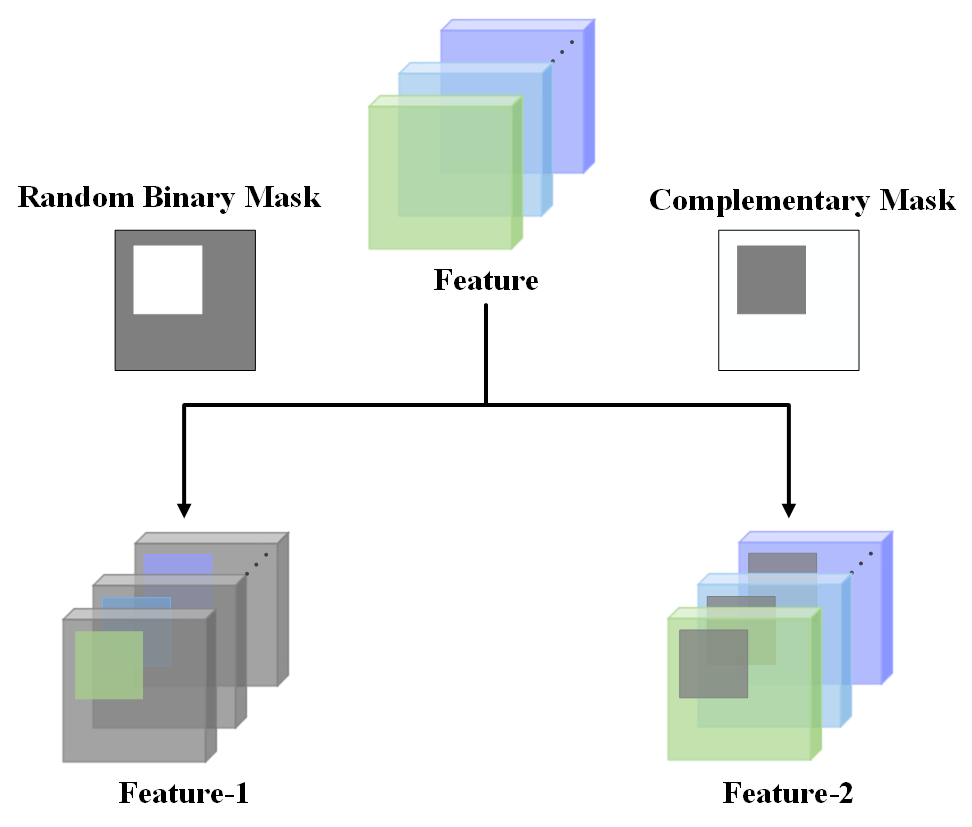}
\caption{Illustration of feature segmentation used in our method. In each iteration, we use a random binary mask to divide the feature into two parts.}
\label{fig1}
\end{center}
\end{figure}

In particular, some methods improve the generalizations by strengthening the learning of local feature. As a representative regularization method, dropout~\cite{srivastava2014dropout} randomly discards parts of internal feature in the network to improve the learning of the remaining feature. Be your own teacher(BYOT)~\cite{zhang2019your} improves the efficiency of network training the by squeezing the knowledge in the deeper portion of the networks into the shallow ones. These approaches are effective to network training because the network is not able to learn the feature sufficiently. In this paper, we focus on developing a new training strategy for CNN for strengthening the network’s learning of local feature.

Our motivation stems from an phenomenon in the training of CNN: information loss will inevitably occur after going through the pooling layer, which leads to the lack of local feature learning in network. We perform several visualization experiments~\cite{zhou2016learning} to explore the difference between global feature and local feature, and results empirically suggest that the semantic between different parts of the feature is different, and local feature contain different semantic compared with global feature, while the network will inevitably lose these information during feedforward propagation. Containing more knowledge of feature is benefit to the training of CNN, but how to make the local feature of images sufficiently mined by network is a big challenge.

Based on the above observation, we propose a parameter-free training method: Feature Mining, which aims to make the network learn local feature more efficiently during training. The Feature Mining approach requires two steps: feature segmentation and feature reusing. During feature segmentation, we divide the feature before pooling layer into two complementary parts by a binary mask as shown in Figure~\ref{fig1}. Feature reusing denotes that the two parts will pass through a global average pooling(GAP) layer and a fully connection layer respectively to generate two independent outputs, and the three cross-entropies will be added to form the final loss functions. In this way, different local feature can participate in the network training, so that the network is able to learn more abundant feature expression. Feature Mining can also be extended to any layers of the network. Specially, adding Feature Mining in the last two layers of the network  not only further improves the performance of the network, but also strengthens the shallow learning ability of network. Due to its inherent simplicity, Feature Mining has a plug-and-play nature. Beside, we do not need to set any hyper parameter when using Feature Mining, which can reduce the operating cost and maintain the stability of the method for performance improvement.

We demonstrate the effectiveness of our simple yet effective training method using different CNNs, such as ResNet~\cite{he2016deep}, DenseNet~\cite{huang2017densely}, Wide ResNet~\cite{zagoruyko2016wide} and PyramidNet~\cite{han2017deep} trained for image classification tasks on various datasets including CIFAR-100~\cite{krizhevsky2009learning}, TinyImageNet and ImageNet~\cite{russakovsky2015imagenet}. Great accuracy boost is obtained by Feature Mining, higher than other dropout~\cite{srivastava2014dropout,tompson2015efficient,ghiasi2018dropblock} and self distillation methods~\cite{zhang2019your,yun2020regularizing}. In addition, our training method has strong compatibility and can be superimposed with the mainstream data augmentation~\cite{zhang2017mixup} and self distillation method. Besides, we verify the wide applicability of Feature Mining in fine-grained classification~\cite{wah2011caltech}, object detection~\cite{everingham2010pascal} and practical scenarios~\cite{cao2019learning,zhang2016understanding}.

In summary, we make the following principle contributions in this paper:
\begin{itemize}
\item We proposed a novel training strategy Feature Mining aims to make the network learn local feature more efficiently. Our method enjoys a plug-and-play nature and is parameter-free.
\item The prosed training approach can obviously improve the network’s performance with the increase of little training cost and can be superimposed with the mainstream training methods.
\item Experiments for four kinds of convolutional neural networks on three kinds of datasets are conducted to prove the generalization of this technique. Tasks in four different scenarios demonstrate the wide applicability of our method.
\end{itemize}

\begin{figure}[t]
\begin{center}
\includegraphics[width=0.95\linewidth]{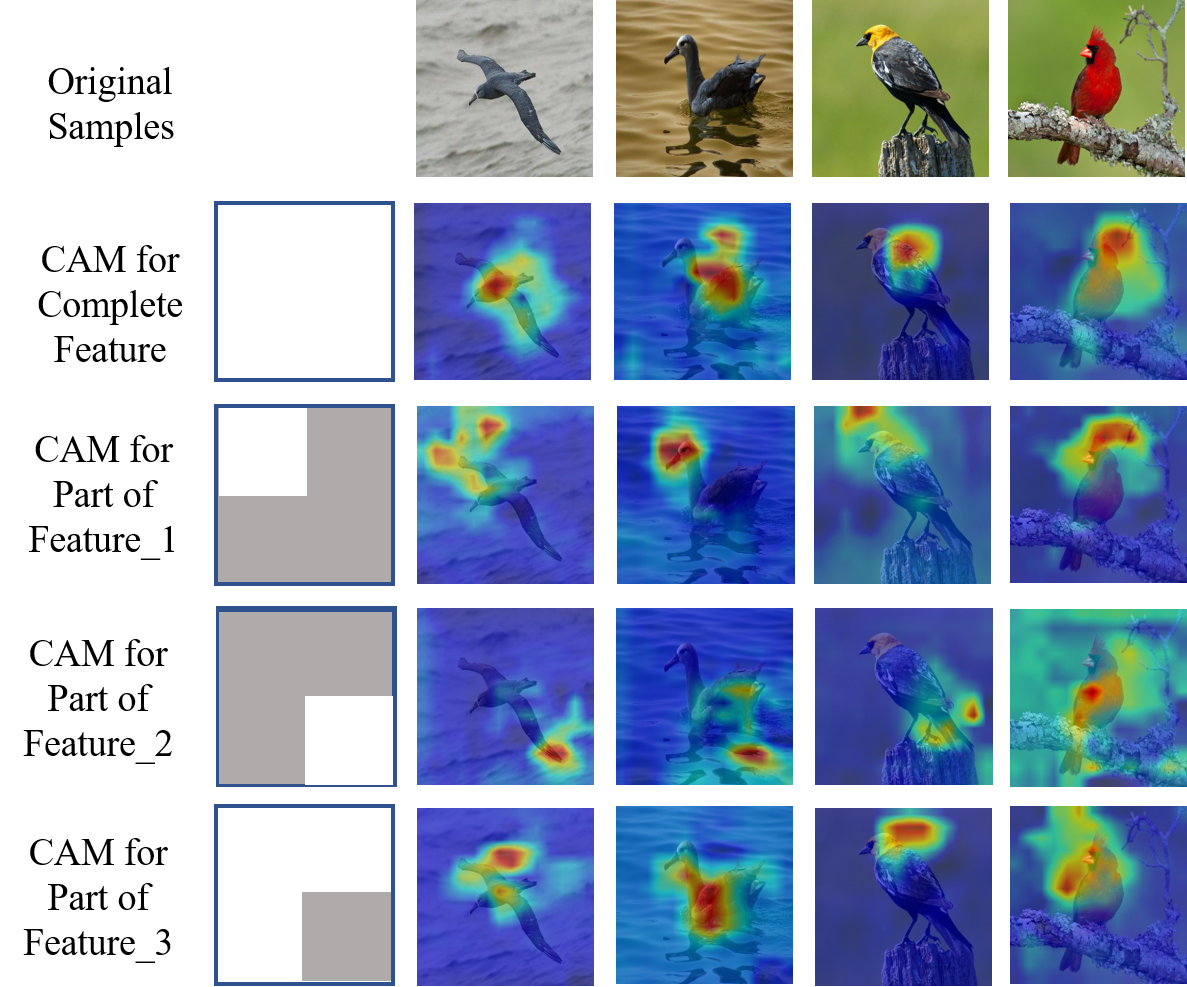}
\caption{Class activation maps (CAM)~\cite{zhou2016learning} for ResNet-50 model trained on CUB-200-2011. Each line represents the visualization results of the same network in different images, while the networks beginning from the third line were trained with a fixed binary masks on the the last layer, which means only part of the features of the last layer of the network could participate in the network training. The first column represents the composition of these binary masks, the gray area represents the discarded area, and only the features in the white area participate in the training }
\label{fig2}
\end{center}
\end{figure}

\begin{figure*}[t]
\begin{center}
\includegraphics[width=0.9\linewidth]{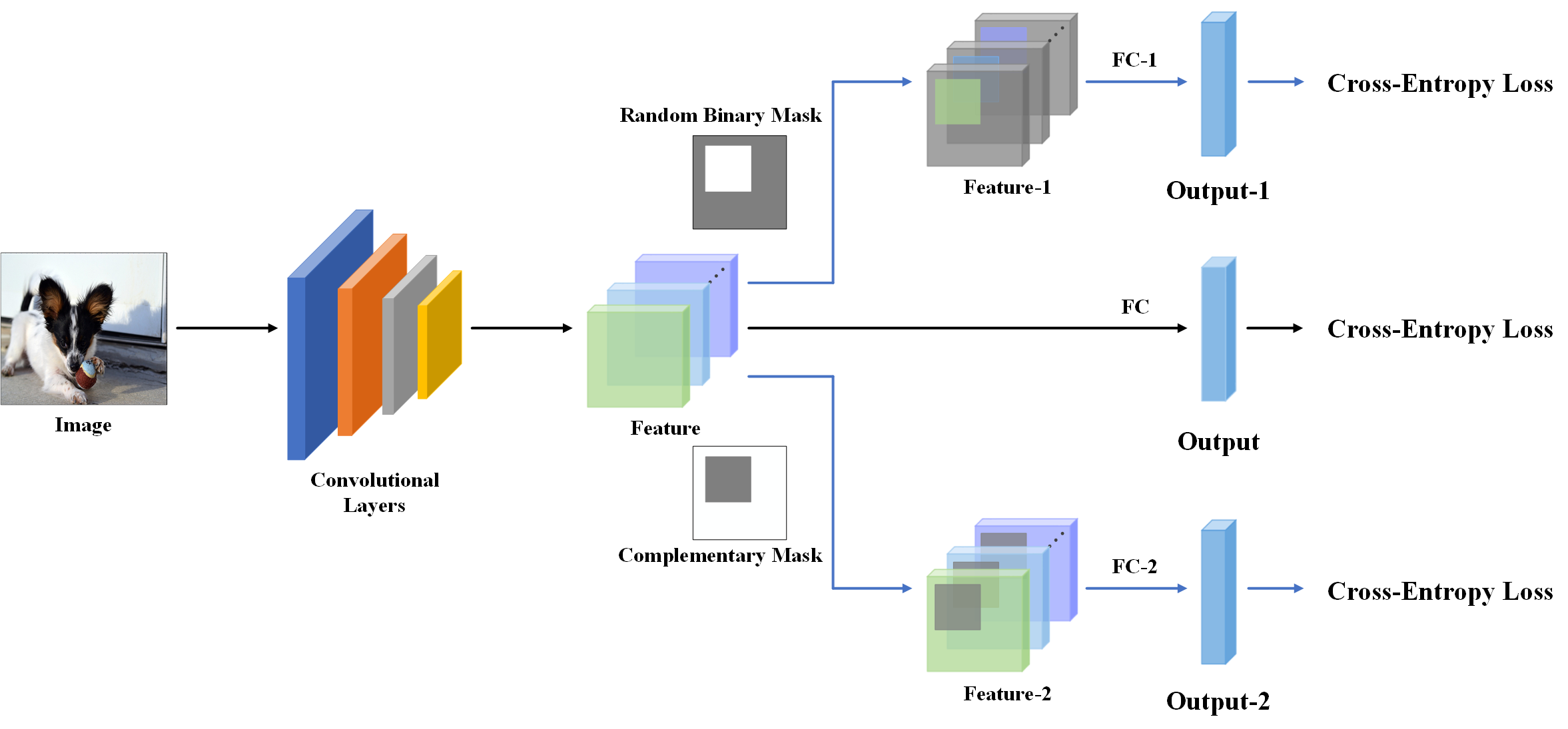}
\caption{This figure shows the details of a CNN training with proposed Feature Mining. There are two stages in Feature Mining: feature segmentation and feature reusing. The feature is divided into two parts and the final three cross-entropies are added as the final loss function.}
\label{fig3}
\end{center}
\end{figure*}

\section{observation}
In order to explore the semantic difference among the local feature, we conducted series of experiments on the visualization of CNN. During the network training, we multiplied the features of the last layer of the network by different fixed binary masks, so that only part of the features of the last layer of the network can participate in the network training. Class activation maps (CAM)~\cite{zhou2016learning} was chosen to visualize the trained model. Specifically, the ResNet50~\cite{han2017deep} trained by CUB-200-2011~\cite{wah2011caltech} was used as the baseline model. All the models were trained from scratch for 200 epochs with batch size 32 and the learning rate is decayed by the factor of 0.1 at epochs 100, 150.

As shown in Figure~\ref{fig2}, the second line shows the visualization results of the complete feature in the last layer, while next lines show the visualization results of differen local feature. We can see that the complete feature generally focused on the global information of the target, while the local feature focused more on the relevant local information according to their locations. Obviously, the image information concerned by local feature is different from that of global feature, and even the semantic information between different local feature is different. However, Due to the feed-forward mechanism of CNN, the network only gets the output depending on the global feature information, which may cause the loss of information contained in the local feature. How to make full use of feature information, especially the local feature, in network training? We will see how to address this issue next.

\section{Feature Mining}

Feature Mining is designed to improve the learning of local feature by networks. Here we present some notations which will be used in the following. For a given training sample $(x, y)$ which $x \in \mathbb R^{W\times H\times C}$ denotes the training image and $y\in Y =\{1,2,3,...,C\}$ denotes the training label, and a network with $N$ classifiers. $X$ denotes the network’s feature of the training sample. For the $n$ classifier, its output is $a^n$. We use softmax to compute the predicted probability:
\begin{equation}
\begin{aligned}
&p_i^n = \frac{exp(a_i^n)}{\sum_{j=1}^n exp(a_j^n)}
\end{aligned}
\end{equation}
Next we will describe the proposed Feature Mining strategy in detail, which consists of two steps $i.e$, feature segmentation and feature reusing, as depicted in Figure~\ref{fig2}.

\subsection{Feature segmentation}
In order to make the network learn different local feature, we first divides high-dimensional feature into two complementary parts by using a binary mask. After obtaining the network’s feature $X$ of the last pooling layer, we generate a binary mask $M$ with a bounding box coordinates $B = (r_x, r_y, r_w, r_h)$ indicating the cropping regions on the binary mask. Like CutMix~\cite{yun2019cutmix}, the box coordinates are uniformly sampled according to:
\begin{equation}
\begin{aligned}
&r_x \sim {\rm Unif}(0,W), r_w = W\sqrt{1-\lambda} \\
&r_y \sim {\rm Unif}(0,H), r_h = H\sqrt{1-\lambda}
\end{aligned}
\end{equation}
making the cropped area ratio $ \frac{r_wr_h}{WH} = 1-\lambda$. $\lambda$ is sampled from the uniform distribution $(0, 1)$. The elements in $B$ are set to one and other elements in $M$ are set to zero. We define the segmentation operation as
\begin{equation}
\begin{aligned}
X_1 &= X \odot M \\
X_2 &= X \odot (1-M)
\end{aligned}
\end{equation}
where $X_1$ and $X_2$ are complementary and contain different knowledge of $X$, which means $X_1$ only contains the feature information within $B$ while $X_2$ only contains the feature information outside $B$. We use block to segment feature based on the characteristics of convolutional neural network: adjacent neurons have similar semantics, and block form segmentation can ensure that the two parts of feature have great differences. Note that this operating can also be applied in any layer to get different representative feature and is not limited by the network structure. For a detailed discussion of the effect of the segmentation method, please refer to Subsection .

\begin{figure*}[t]
\begin{center}
\includegraphics[width=0.85\linewidth]{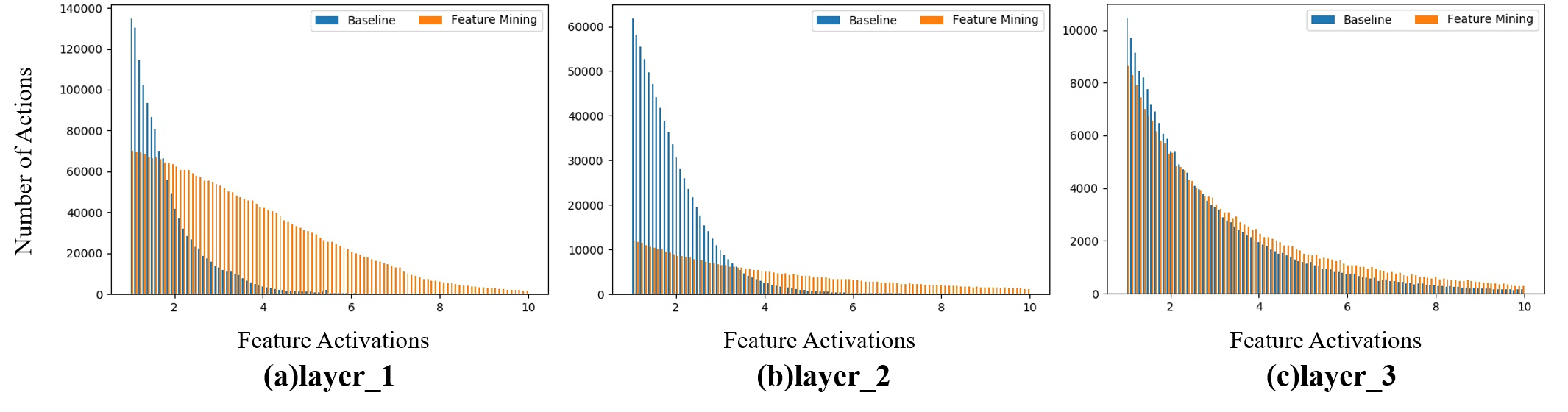}
\caption{Number of feature activations averaged over all 64 random samples. A standard ResNet-110 is compared with a ResNet-110 trained with Feature Mining at three different layers.}
\label{fig4}
\end{center}
\end{figure*}

\subsection{Feature reusing}
After feature segmentation, we get a set of extra complementary feature $X_1$ and $X_2$. Based on the observation in section 2, the semantics of $X$, $X_1$ and $X_2$ are different from each other. In order to make the network learn the information of these three parts of knowledge, we add two groups of global average pooling layers and fully connection layers on the basis of the original network. $X_1$ and $X_2$ will enter their own global average pooling layer and fully connection layer respectively to get the new outputs $a^1$ and $a^2$. The probabilities $p^1$ and $p^2$ corresponding to these two outputs  will also be calculated with the real label of the image to get two new cross-entropies. We add these two cross-entropies with the cross-entropy of $X$ to get the final loss function.
\begin{equation}
\begin{aligned}
loss = \sum_{i=0}^n CrossEntropy(p^i, y)
\end{aligned}
\end{equation}
These three independent cross-entropies work together on the back propagation of the network based on the loss function, and affect the gradient update of the network parameters to make the neural network learn the knowledge information of the overall feature and local feature simultaneously, so as to achieve the purpose of feature reusing.

\subsection{Application}
Feature Mining can be theoretically plugged behind any convolutional layer within CNN, and can be superimposed on multiple layers at once training. In practice, we choose the last one or two layer for our method. Note that the binary masks used as segmentation are generated randomly in each network iteration, which makes the network continuously learn the different complementary parts of feature, so that the training of the whole network is more efficient and robust. To stress again, as a lightweight parameter-free strategy, Feature Mining is added to these networks only in training phase and the network in test phase is unchanged.

\subsection{Why Does Feature Mining Help?}
In order to verify whether Feature Mining enhances the network's ability of learning feature, we quatitatively compared the average magnitude of feature activations in ResNet-110 on CIFAR-100 when trained with and without Feature Mining. Note our method are used in the last two layers. The training details can be seen in Subsection 4.1. Feature activations over 64 randomly sampled images are quantified. Besides, considering the predominant low values of activations resulting from the sparsity in deep learning model, only the activations greater than 0.5 are retained for better illustration. The results are displayed in Figure 4. We observed that the model trained with our method had more high-values activations than the baseline, especially in the first layer, which illustrates that Feature Mining can make network learn a wider range of activation value. It also proves that enhancing the learning of local feature can make the network capture more representative features.


\section{Experiments}
\begin{table*}[t]
\begin{center}
\begin{tabular}{lcccccc}
\toprule[1.2pt]
\midrule

Method     &ResNet-56   &ResNet-110 &ResNet-164    &DenseNet  &WRN-28-10 &PyramidNet\\
\midrule

Baseline     &73.71$\pm$0.28     &74.71$\pm$0.23   &74.98$\pm$0.23        &77.25$\pm$0.20       &79.35$\pm$0.20   &81.54$\pm$0.31\\

Dropout     &73.81$\pm$0.27     &74.69$\pm$0.33   &75.15$\pm$0.11     &77.45$\pm$0.12       &79.21$\pm$0.15   &81.31$\pm$0.24\\

SpatialDropout       &74.38$\pm$0.17       &74.76$\pm$0.12    &75.34$\pm$0.09     &77.97$\pm$0.16      &79.63$\pm$0.09  &81.60$\pm$0.17\\

DropBlock    &73.92$\pm$0.10     &74.92$\pm$0.07 &75.51$\pm$0.14     &77.33$\pm$0.09            &79.44$\pm$0.24    &81.57$\pm$0.21\\
\midrule

FM(1layer)   &{\bfseries 74.55$\pm$0.11}     & {\bfseries 75.79$\pm$0.08}     &{\bfseries 76.51$\pm$0.13}      &{\bfseries 78.24$\pm$0.12}  &{\bfseries 79.75$\pm$0.16}   &{\bfseries 82.13$\pm$0.25}\\

FM(2layers)   &{\bfseries 75.35$\pm$0.07}     & {\bfseries 76.90$\pm$0.13}     &{\bfseries 78.31$\pm$0.21}      &{\bfseries 78.78$\pm$0.09}  &{\bfseries 80.90$\pm$0.22}  &{\bfseries 82.87$\pm$0.15}\\

\midrule
\bottomrule[1.2pt]
\end{tabular}
\end{center}
\caption{Validation accuracy (\%) on CIFAR100 dataset using CNN architectures of ResNet-56, ResNet-110, ResNet-164, Densenet-100, WRN-28-10, PyramidNet-110-270 (Top accuracy are in bold). }
\label{tab1}
\end{table*}

In this section, we investigate the effectiveness of Feature Mining for multiple computer vision tasks. We first conducted extensive experiments on image classification(Section 4.1) and fine-grained image classification(Section 4.2). Next, we studied the effect of Feature Mining on object detection(Section 4.3). Besides, we evaluated the performance of our method in practical scenarios(Section 4.4). We also show the ablation study of Feature Mining in Section 4.5. All experiments are performed with Pytorch~\cite{paszke2017automatic} on Tesla M40 GPUs. The highest validation accuracy over the full training course is chosen as the result. If not specified, all results reported are averaged over 4 runs.

\subsection{Image Classification}
\subsubsection{CIFAR Classification}
The CIFAR100 dataset~\cite{krizhevsky2009learning} consists of 60,000 32$\times$32 color images of 100 classes, each with 600 images including 500 training images and 100 test images. For ResNet~\cite{he2016deep} and Wide ResNet~\cite{zagoruyko2016wide}, the mini-batch size was set to 128 and the models were trained for 300 epochs with initial learning rate 0.1 decayed by factor 0.1 at epochs 150, and 225. We changed the batch size to 64 when training with DenseNet~\cite{huang2017densely}. As mentioned in~\cite{han2017deep} , we changed the initial learning rate to 0.5 and decreased it by a factor of 0.1 at 150 and 225 epochs when training PyramidNet~\cite{han2017deep} .

\noindent \textbf{Comparison with dropout methods.} We adopt ResNet-56, ResNet-110, ResNet-164, DenseNet-100-12, Wide ResNet-28-10 and PyramidNet-110-270 as the baselines to evaluate Feature Mining’s generalization for different layers and structures of networks. We test Feature Mining with the last layer and last two layers. Dropblock~\cite{ghiasi2018dropblock} is applied to the output of first two groups. Dropout~\cite{srivastava2014dropout} and SpatialDropout~\cite{tompson2015efficient} are applied to the output of penultimate group. Table~\ref{tab1} shows that Feature Mining outperforms other baselines consistently. In particular, Feature Mining improves the top-1 accuracy of the cross-entropy loss from 74.98\% to 78.31\% of ResNet-164 under the CIFAR100 dataset.

Our method only used in the last layer is better than the performance of the three dropouts, we consider the reason is that Dropout and its variant methods strengthen the part of neurons' learning by discarding other part of neurons, which will inevitably cause the information loss during training phase. While Feature Mining can make the network simultaneously learn the different regions of feature without information loss, so that the network can achieve more desirable performance. We can also find that the network performance of Feature Mining used in the last two layers is much better than that only used in the last layer, which indicates that the local feature in the front layer is also benefit to the network training.

\noindent \textbf{Comparison with self distillation methods.}  We also compare our method with recent proposed self distillation techniques such as Be Your Own Teacher(BYOT)~\cite{zhang2019your} and Class-wise knowledge distillation(CS-KD)~\cite{yun2020regularizing}. BYOT improves the performance of the network by making the shallow network learn the last layer of knowledge, we take the output of the last layer in ResNet110 as the teachers of all the front layers, and provide the final results after ensemble for fair comparision. CS-KD improves the generalization of the same kind of images from the perspective of intra class distillation.

As shown in Table~\ref{tab2}, Feature Mining shows better top-1 accuracy on ResNet-110 compared with BYOT and CS-KD. Note that BYOT adds extra overhead in training, while the time of our method is negligible. Besides, we combine Feature Mining with these two methods and improve improves the top-1 accuracy of ResNet-110 from 76.68\% to 77.34\% of BYOT,  and from 75.92\% to 77.68\% of CS-KD. It shows that our method does not conflict with these methods, even in the case of improving the performance of the front layer and reducing the gap within the class, the network still needs to strengthen the learning of local feature.
\begin{table}[h]
\begin{center}
\begin{tabular}{lr}
\toprule[1.2pt]
\midrule

Method    &Validation Accuracy(\%)  \\
\midrule
ResNet-110 &74.71$\pm$0.23\\

ResNet-110+BYOT &76.68$\pm$0.14\\

ResNet-110+CS-KD &75.92$\pm$0.08\\

ResNet-110+FM &76.90$\pm$0.21\\

ResNet-110+FM+BYOT &77.34$\pm$0.16\\

ResNet-110+FM+CS-KD &{\bfseries 77.68$\pm$0.11}\\

\midrule
\bottomrule[1.2pt]
\end{tabular}
\end{center}
\caption{
The performance comparison with self distillation methods on CIFAR100 dataset.
}
\label{tab2}
\end{table}

\noindent \textbf{Comparison with data augmentation method.} We investigate orthogonal usage with other types of regularization methods such as Mixup~\cite{zhang2017mixup}. Mixup utilizes convex combinations of input pairs and corresponding label pairs for training. We combine our method with Mixup regularization by combining the input and label processed by Mixup with ResNet-110 which using Feature Mining on the last two layers. Table~\ref{tab3} shows the effectiveness of our method combined with Mixup regularization. Interestingly, this simple idea significantly improves the performances of classification task. In particular, our method improves the top-1 accuracy of Mixup regularization from 79.22\% to 81.04\% on ResNet-164, that proves our method is compatible with the data augmentation method.

\begin{table}[h]
\begin{center}
\begin{tabular}{lr}
\toprule[1.2pt]

\midrule
Metod    &Validation Accuracy(\%)  \\
\midrule
ResNet-110 &74.71$\pm$0.23\\

ResNet-110+Mixup &77.78$\pm$0.25\\

ResNet-110+FM &76.90$\pm$0.12\\

ResNet-110+FM+Mixup &{\bfseries79.13$\pm$0.17}\\

\midrule

ResNet-164 &74.98$\pm$0.23\\

ResNet-164+Mixup &79.22$\pm$0.12\\

ResNet-164+FM &78.31$\pm$0.09\\

ResNet-164+FM+Mixup &{\bfseries81.04$\pm$0.20}\\

\midrule
\bottomrule[1.2pt]
\end{tabular}
\end{center}
\caption{
The performance comparison with Mixup on CIFAR100 dataset.
}
\label{tab3}
\end{table}

\subsubsection{Tiny ImageNet Classification}
Tiny ImageNet dataset is a subset of the ImageNet~\cite{russakovsky2015imagenet} dataset with 200 classes. Each class has 500 training images, 50 validation images, and 50 test images. All images are with 64$\times$64 resolution. The test-set label is not publicly available, so we use the validation-set as a test-set for all the experiments on Tiny ImageNet following the common practice. The results are summarized on Table~\ref{tab4}. Feature Mining achieves the best performance \% on Tiny ImageNet compared with dropout methods, BYOT and CS-KD. This proves that our method is also generalized for datasets with different data sizes. These results also show the wide applicability of our method, compatible to use with other regularization and self distillation methods.

\begin{table}[h]
\begin{center}
\begin{tabular}{lr}
\toprule[1.2pt]
\midrule
Method    &Validation Accuracy(\%)  \\
\midrule
ResNet110 &62.42$\pm$0.25\\

ResNet110+Dropout &62.32$\pm$0.05\\

ResNet110+Spatial Dropout &62.55$\pm$0.10\\

ResNet110+DropBlock &63.13$\pm$0.29\\

ResNet110+BYOT &63.03$\pm$0.17\\

ResNet110+CS-KD &64.29$\pm$0.11\\

ResNet110+FM &{\bfseries 65.22$\pm$0.06}\\
\midrule
\bottomrule[1.2pt]
\end{tabular}
\end{center}
\caption{
The performance comparison on Tiny ImageNet dataset. The best accuracy is achieved by Feature Mining.
}
\label{tab4}
\end{table}
\subsubsection{ImageNet Classification}
ImageNet-1K~\cite{russakovsky2015imagenet} contains 1.2M training images and 50K validation images labeled with 1K categories. To verify the scalability of our method, we have evaluated our method on the ImageNet dataset with ResNet-50. The model is trained from scratch for 300 epochs with batch size 256 and the learning rate is decayed by the factor of 0.1 at epochs 75, 150, 225. As reported in Table~\ref{tab5}, our method improves 0.94\% of the top-1 accuracy compared with baseline. This shows that our training strategy is also effective for the image recognition of large scale dataset.

\begin{table}[h]
\begin{center}
\begin{tabular}{lr}
\toprule[1.2pt]
\midrule
Method    &Validation Accuracy(\%)  \\
\midrule
ResNet-50 &76.32$\pm$0.02\\

ResNet-50+FM &{\bfseries 77.26$\pm$0.04}\\

\midrule
\bottomrule[1.2pt]
\end{tabular}
\end{center}
\caption{
The performance comparison on ImageNet dataset on ResNet-50.
}
\label{tab5}
\end{table}
\subsection{Fine-grained Image Classification}
The fine-grained image classification aims to recognize similar subcategories of objects under the same basic-level category. The difference of fine-grained recognition compared with general category recognition is that fine-grained subcategories often share the same parts and usually can only be distinguished by the subtle differences in texture and color properties of these parts. CUB-200-2011~\cite{wah2011caltech} is a widely-used fine-grained dataset which consists of images in 200 bird species. There are about 30 images for training for each class. We use ResNet-50 to test the performance of our method on CUB-200-2011 to verify the generalization of different types of computer vision tasks. For fair comparison, the model is trained from scratch for 300 epochs with batch size 32 and the learning rate is decayed by the factor of 0.1 at epochs 150,225.

\begin{figure}[h]
\begin{center}
\includegraphics[width=0.8\linewidth]{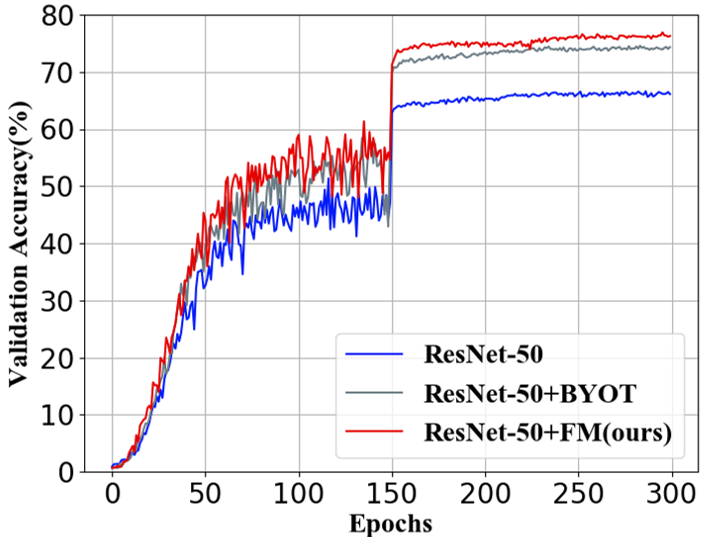}
\caption{Validation accuracy comparison among baseline, BYOT and Feature Mining on CUB-200-2011 with ResNet-50.}
\label{fig5}
\end{center}
\end{figure}

\begin{table}[h]
\begin{center}
\begin{tabular}{lr}
\toprule[1.2pt]
\midrule
Method    &Validation Accuracy(\%)  \\
\midrule
ResNet-50 &66.73$\pm$0.34\\

ResNet-50+BYOT &74.60$\pm$0.41\\

ResNet-50+CS-KD &74.81$\pm$0.23\\

ResNet-50+FM & {\bfseries76.26$\pm$0.27}\\

\midrule
\bottomrule[1.2pt]
\end{tabular}
\end{center}
\caption{
The performance comparison with self distillation on CUB-200-2011 dataset.
}
\label{tab6}
\end{table}
As shown in Table~\ref{tab6}, with Feature Mining, we improve the accuracy of ResNet-50 from 66.73\% to 76.26\%(+9.53\%), which surpasses previous self distillation methods significantly. It also shows that fully utilizing feature information has a high gain effect even on fine-grained image classification which requires more subtle differences. Figure~\ref{fig5} shows the validation accuracy comparison among baseline, BYOT and Feature Mining on CUB-200-2011 with ResNet-50. The accuracy of BYOT and our method is much higher than baseline and the accuracy of Feature Mining is higher than that of BYOT.

\subsection{Object Detection}
In this subsection, we show Thumbnail can also be applied for training object detector in Pascal VOC dataset~\cite{everingham2010pascal}. We use RetinaNet~\cite{lin2017focal} framework composed of a backbone network and two task-specific subnetworks for the experiments. The ResNet-50 backbone which is responsible for computing a convolutional feature map over an entire input image is initialized with ImageNet-pretrained model and then fine-tuned on Pascal VOC 2007 and 2012 trainval data. Models are evaluated on VOC 2007 test data using the mAP metric. We follow the fine-tuning strategy of the original method.

\begin{table}[h]
\begin{center}
\begin{tabular}{lr}
\toprule[1.2pt]
\midrule
Method    &Validation Accuracy(\%)  \\
\midrule
RetinaNet &70.14$\pm$0.17\\

RetinaNet+FM pre-trained &{\bfseries 71.11$\pm$0.14}\\

\midrule
\bottomrule[1.2pt]
\end{tabular}
\end{center}
\caption{
The performance comparison on Pascal VOC dataset.
}
\label{tab7}
\end{table}
As shown in Table~\ref{tab7}, the model pre-trained with Feature Mining achieves the better accuracy (71.11\%), +0.97\% higher than the baseline performance. The results suggest that the model trained with Feature Mining can better capture the target objects, and our methd is an effective training approach for object detection.

\subsection{Practical Scenarios}
\noindent \textbf{A. Label Noise Problem}
In practical scenarios, the dataset is full of error labels because of the expensive and time-consuming labeling process, which seriously affects the performance of the network. This phenomenon is known as label noise problem.
We set up experiments with noisy datasets to see how well Feature Mining performs for different types and amounts of label noise.

Following~\cite{cao2019learning,zhou2020bbn}, we corrupted these datasets manually on CIFAR-100. Two types of noisy label are considered: (1) Symmetry flipping: each label is set to an incorrect value uniformly with probability; (2)Pair flipping: labelers may only make mistakes within very similar classes. We use $\epsilon$ to denote the noise rate.

\begin{table}[h]
\begin{center}
\begin{tabular}{ccll}
  \toprule[1.2pt]
  \midrule
\multicolumn{1}{l}{Noise Type}                                            & \multicolumn{1}{l}{$\epsilon$}                & Method                            & Accuracy(\%) \\
 \midrule
\multirow{4}{*}{\begin{tabular}[c]{@{}c@{}}Symmetric\\ Filp\end{tabular}} & \multirow{2}{*}{0.5}                     & ResNet-110                        &45.73          \\
                                                                          &                                          & ResNet-110+FM                     &53.31(+7.58)   \\
                                                                          \cmidrule{2-4}
                                                                          & \multirow{2}{*}{0.2}                     & ResNet-110                        &61.16          \\
                                                                          &                                          & ResNet-110+FM                     &68.12(+6.96)          \\
\midrule
\multirow{4}{*}{\begin{tabular}[c]{@{}c@{}}Pair\\ Filp\end{tabular}}      & \multirow{2}{*}{0.5}                     & ResNet-110                        &35.73          \\
                                                                          &                                          & ResNet-110+FM                     &38.78(+3.05)          \\
                                                                          \cmidrule{2-4}
                                                                          & \multirow{2}{*}{0.2}                     & ResNet-110                        &63.83          \\
                                                                          &                                          & ResNet-110+FM                     &71.22(+7.39)          \\
  \midrule
  \bottomrule[1.2pt]
\end{tabular}
\end{center}
\caption{Average test accuracy on noisy CIFAR-100 using ResNet-110 with and without Feature Mining. }
\label{tab8}
\end{table}

Table~\ref{tab8} compares the results with and without Feature Mining using ResNet-110. Our method has all made huge improvements on different types and amounts of noise compared with baseline (which only uses cross-entropy to classify). One possible implication of this is that our method can make the convolutional layers training more sufficient, and the robust network has stronger anti noise ability. Together these results provide strong proofs that Feature Mining helps raise the internal noise tolerance of the network.

\noindent \textbf{B. Small Sample Size Problem} In some real industrial scenarios, the size of dataset may be small, which would bring difficulty to the network training. Besides, \cite{bousquet2002stability} denoted that reducing the size of the training dataset can measure the generalization performance of models. Therefore, we evaluate the effectiveness of Feature Mining on small sample size problem along with varying the size of the training data.

\begin{figure}[h]
\begin{center}
\includegraphics[width=0.90\linewidth]{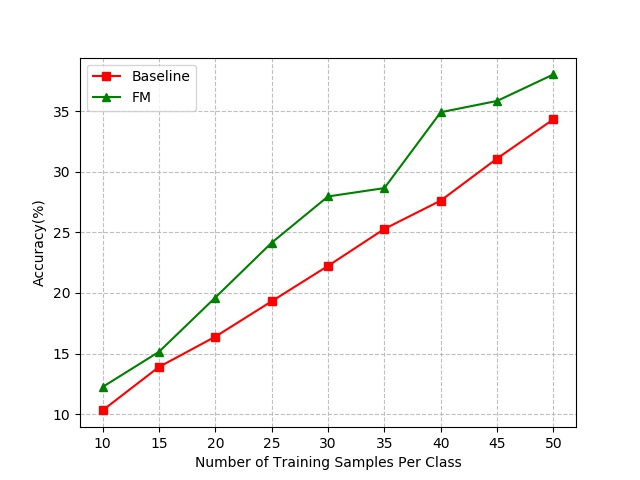}
\caption{Accuracy(\%) of ResNet-56 with and without Feature under varying number of training samples per class in CIFAR-100.}
\label{fig6}
\end{center}
\end{figure}

We constructed a series of sub-datasets of CIFAR-100 via randomly choosing samples for each class. Figure~\ref{fig6} compares the results obtained from ResNet-56 trained with and without our strategy. Feature Mining is used on the last two layers of network. From the graph, we can see that our method yields much higher accuracies compared to the vanilla setting on all of the sub-datasets, in accordance with the expectations. The performance of Feature Mining proves that our method is able to improve the generalization of network effectively.

\subsection{Ablation Study}

\noindent \textbf{A. Influence of different choices of Feature Mining on network performance}
We explore different design choices for Feature Mining, these choices are classified into two groups.

\begin{table}[h]
\begin{center}
\begin{tabular}{lr}
\toprule[1.2pt]
\midrule
Method    &Accuracy(\%)  \\
\midrule
ResNet56 &73.71$\pm$0.28\\

ResNet56+FM(1layer) &74.55$\pm$0.11\\

ResNet56+FM(2layers) &{\bfseries 75.35$\pm$0.07}\\

ResNet56+FM(3layers) &75.29$\pm$0.12\\

\midrule

ResNet56+Dropout-FM &74.12$\pm$0.23\\

ResNet56+SpatialDropout-FM &74.01$\pm$0.17\\

ResNet56+non-complementary FM &74.23$\pm$0.09\\

\midrule
\bottomrule[1.2pt]
\end{tabular}
\end{center}
\caption{The effect of different choices of Feature Mining on network performance.}
\label{tab9}
\end{table}

As shown in Table~\ref{tab9}, the first group discusses the number of layers that use Feature Mining. ResNet-56 contains three layers, we test the Feature Mining on the last layers, last two layers and whole three layers. All three models outperform the baseline model from 74.55\% to 75.35\%, while the model with last two layers using Feature Mining  perform better than others. It shows that the penultimate layer's local feature is benefit to the network training, and Feature Mining is an effective approach reusing these feature. While the feature of the first layer are not yet mature, adding learning of the feature in the first layer can not help the network training better.

The second group focuses on the feature segmentation approach. The feature segmentation determines the additional feature pattern of the network learning, which plays a crucial role in the performance of our method. Dropout-FM denotes the binary mask is set by random points like dropout. SpatialDropout-FM denotes the feature is divided by channels like spatial dropout. The performance of these segmentation can not compete with Feature Mining, which proves the segmentation feature based on region can ensure the higher difference between the two parts and make the network learning more abundant. Non-complementary FM denotes we randomly choose two parts of feature instead of complementary ones, its performance is still not as good as the complementary way, that verifies the difference between the two parts of feature is important again.

\noindent \textbf{B. Analysis of time and memory overhead }
Table 10 provides some complexity statistics of Feature Mining, which indicates that the computation cost introduced by Feature Mining is negligible, especially when our method is only used in the last layer. Besides, these additional cost is introduced only in training phase and the inference cost is not increased at all.

\begin{table}[h]
\begin{tabular}{lcl}
\toprule[1.2pt]
\midrule
Method    & \begin{tabular}[c]{@{}c@{}}GPU memory\\ (MB)\end{tabular} & \multicolumn{1}{c}{\begin{tabular}[c]{@{}c@{}}Training time\\ (sec/iter)\end{tabular}} \\
\midrule
ResNet-56            &1740                                                             &0.094                                                                                        \\
+FM(1layer) &1773                                                               &0.098                                                                                        \\
+FM(2layers) &1959                                                               &0.113                                                                                      \\
\midrule
\bottomrule[1.2pt]
\label{tab10}
\end{tabular}
\caption{
Computation cost introduced by Feature Mining. The statistics are obtained on a Tesla M40 GPU with batch size = 128. The training time is averaged over the first 300 iterations.
}
\end{table}

\section{Related Work}
\subsection{Regularization Methods}
The regularization methods aim to improve the generalization of neural networks. Dropout~\cite{srivastava2014dropout} injects noise into feature space by randomly zeroing the activation function to avoid overfitting. \cite{wan2013regularization,tompson2015efficient,ghiasi2018dropblock,keshari2019guided,ouyang2019attentiondrop,hou2019weighted} are also proposed as variants of Dropout. Besides, Batch Normalization~\cite{ioffe2015batch} improves the gradient propagation through network by normalizing the input for each layer.

\subsection{Data Augmentation}
Data augmentation is a effective method for network training. Random cropping, horizontal flipping~\cite{krizhevsky2012imagenet} are the most commonly used data augmentation techniques. By randomly removing contiguous sections of input images, Cutout~\cite{devries2017improved} improves the robustness of network. Random Erasing~\cite{zhong2020random} randomly selects a rectangle region in an image and erases its pixels with random values. Input mixup~\cite{zhang2017mixup} creates virtual training examples by linearly interpolating two input data and corresponding one-hot labels. Inspired by Cutout and Mixup, CutMix~\cite{yun2019cutmix} cut patches and pasted among training images.

\subsection{Knowledge Distillation}
Knowledge distillation~\cite{hinton2015distilling} is proposed for model compression. Unlike traditional knowledge distillation methods that require teacher and student models, self distillation distills knowledge itself. Data-distortion~\cite{xu2019data} transfers knowledge between different augmented versions of the same training data. Be Your Own Teacher~\cite{zhang2019your} improves the efficiency of network training the by squeezing the knowledge in the deeper portion of the networks into the shallow ones. Class-wise self knowledge distillation~\cite{yun2020regularizing} improves the generalization of the same kind of images from the perspective of intra class distillation.

\section{Conclusion}
We have proposed a novel training strategy named Feature Mining which strengthens the learning of local feature during training period. Through feature segmentation and feature reusing, our method can make the network better capture different feature, so as to improve the efficiency of training. Extensive experiments prove that Feature Mining brings stable improvement to different classification datasets on various models, and can be applied to various tasks, including fine-grained image classification, object detection, label noise, and small data regime. For future work, we plan to find better segmentation strategies  for Feature Mining by using reinforcement learning and apply our method to more types of visual tasks.

\bibliographystyle{ACM-Reference-Format}
\bibliography{re}

\end{document}